\documentclass[runningheads]{llncs}
\usepackage[T1]{fontenc}
\usepackage{graphicx}
\usepackage{amsfonts}
\usepackage{multirow}
\usepackage{amsmath}
\usepackage{algorithm}
\usepackage{algorithmicx}
\usepackage{algpseudocode}
\usepackage{enumitem}

\newcommand{\E}{\mathbb{E}}
\DeclareMathOperator*{\regularization}{\ensuremath{{\theta}}}
\newcommand{\set}[1]{\mathcal{#1}}
\newcommand{\pnorm}[1]{\lVert{#1}\rVert}
\DeclareMathOperator*{\loss}{{\ell}}
\newcommand{\RN}{\mathbb{R}}
\newcommand{\x}{\ensuremath{\vec{x}}}

\newcommand{\y}{\ensuremath{y}}
\newcommand{\xorig}{\ensuremath{\vec{x}_{\text{orig}}}}

\newcommand{\setX}{\ensuremath{\set{X}}}
\newcommand{\setS}{\ensuremath{\set{S}}}
\newcommand{\setY}{\ensuremath{\set{Y}}}
\newcommand{\xcf}{\ensuremath{\vec{x}_{\text{cf}}}}
\newcommand{\ycf}{\ensuremath{y_{\text{cf}}}}

\newcommand{\deltacf}{\ensuremath{\vec{\delta}_\text{cf}}}
\newcommand{\w}{\ensuremath{\vec{w}}}

\newcommand{\myCF}[2]{\ensuremath{\text{CF}(#1,#2)}}
\newcommand{\dimsym}{d}
\newcommand{\setD}{\set{D}}

\newcommand{\classifier}{\ensuremath{h}}
\newcommand{\refeq}[1]{Eq.~\eqref{#1}}
\newcommand{\refdef}[1]{Definition~\ref{#1}}

\newcommand{\expl}{\ensuremath{z}}
\newcommand{\setExpl}{\ensuremath{\set{E}}}

\begin{document}
%
%\title{Tracing Explanations back to Training Samples}
\title{Towards Understanding the Influence of Training Samples on Explanations}

\author{Andr\'e Artelt\inst{1,2}\orcidID{0000-0002-2426-3126} \and
Barbara Hammer\inst{1}\orcidID{0000-0002-0935-5591}}
\authorrunning{A. Artelt et al.}
\institute{Bielefeld Universty, Germany \and
University of Cyprus, Cyprus\\
\email{\{aartelt,bhammer\}@techfak.uni-bielefeld.de}}
\maketitle              % typeset the header of the contribution
\begin{abstract}
Explainable AI (XAI) is widely used to analyze AI systems' decision-making, such as providing counterfactual explanations for recourse. When unexpected explanations occur, users may want to understand the training data properties shaping them.
Under the umbrella of data valuation, first approaches have been proposed that estimate the influence of data samples on a given model. This process not only helps determine the data's value, 
but also offers insights into how individual, potentially noisy, or misleading examples affect a model, which is crucial for interpretable AI. In this work, we apply the concept of data valuation to the significant area of model evaluations, focusing on how individual training samples impact a model's internal reasoning rather than the predictive performance only.
Hence, we introduce the novel problem of identifying training samples shaping a given explanation or related quantity, and investigate the particular case of the cost of computational recourse. We propose an algorithm to identify such influential samples and conduct extensive empirical evaluations in two case studies.

\end{abstract}

\section{Introduction}
Today, numerous AI and ML systems are deployed in real-world applications~\cite{zhao2023survey,ho2022cascaded},
demonstrating impressive performance yet remaining imperfect. Issues like failures, fairness concerns, and vulnerabilities to manipulations such as data poisoning can pose risks. Therefore, transparency is crucial to prevent failures, build trust, and ensure safe deployment. Policymakers have recognized this, embedding transparency in regulations like the EU's GDPR~\cite{GDPR}  and the EU AI Act~\cite{euAiAct21}. Explanations are a key way to achieve transparency, shaping the field of Explainable AI (XAI)~\cite{dwivedi2023explainable}. Due to diverse use cases and users, many explanation methods exist, including popular ones like LIME~\cite{Lime}, SHAP~\cite{sundararajan2020many}, and counterfactual explanations~\cite{CounterfactualWachter}, that offer computational recourse.

Although current XAI methods can reveal the internal logic of a model, they do not clarify the reasons behind this logic. However, when faced with surprising or undesirable explanations, like an expensive or impractical recourse recommendation, users might want an "explanation of the explanation." This need is also heightened by recent findings that explanations can be manipulated or poisoned~\cite{artelt2024effect,baniecki2023adversarial,baniecki2022fooling}, undermining users' trust. A possible way to address this is by tracing explanations back to influential training samples, which shaped the model's internal logic. The knowledge of such influential training samples could not only reveal insights into the relevance of certain training samples for shaping the model's reasoning, but also allow for sanity checks, removal, or correction of the training data if necessary. \emph{Such an approach would explain a model's reasoning within the training data space, complementing traditional XAI methods that focus on the feature or model parameter space.} To the best of our knowledge, no prior work has looked into this aspect.

\paragraph{Our contributions:}
We introduce and formalize the novel problem of \emph{analyzing the influence of training samples on explanations} and propose an algorithm for identifying such influential training samples.
We focus on two specific cases involving counterfactual explanations for computational recourse derived from the trained model: 1) Identifying training samples that strongly influence the average cost of recourse, and 2) Identifying samples that significantly affect the cost difference of recourse between two protected groups, indicating a group fairness violation. We conduct extensive empirical evaluations of our proposed algorithms and compare them with baseline approaches.

\section{Foundations}
\subsection{Data-Valuation}\label{sec:foundations:datashap}
Data valuation~\cite{sim2022data} focuses on assessing the importance of individual training samples on predictive performance, quantifying each sample's contribution to the final model. This knowledge can be used to compensate users for their data, verify the accuracy of highly relevant samples, or acquire more data similar to the most significant training samples.

In this context, the Data-SHAP method~\cite{ghorbani2019data} constitutes a popular model-agnostic method that carries the concept of Shapley-Values~\cite{sundararajan2020many} over to data valuation.
Here, the Shapley value $\phi_i\in\RN$ states the contribution of the $i$-th player (e.g. feature or training data point) to some quantity of interest $V:\setS\mapsto\RN$ (also called value function) of a predictive function $\classifier:\setX\to\setY$ derived from a given training data set $\setS$.
In data valuation, as stated before, the property of interest is usually the predictive performance of $\classifier(\cdot)$, e.g.\ the accuracy is used as an implementation of the value function $V(\cdot)$.
Requiring some equitable properties, it can be shown~\cite{ghorbani2019data} that the solution of $\phi_i$ is given as:
\begin{equation}\label{eq:data_shape}
    \phi_i = C\sum_{\setS\subseteq\setD-\{i\}}\frac{V(\setS\cup\{i\}) - V(\setS)}{\binom{|\setD|-1}{|\set{S}|}}
\end{equation}
where $C$ is a constant, and $\{i\}$ refers to the $i$-th sample in a given set $\setD$.
Like Shapley-Values, the computation of~\refeq{eq:data_shape} is computationally infeasible. Therefore, in~\cite{ghorbani2019data} a Monte-Carlo approximation of~\refeq{eq:data_shape} is proposed:
\begin{equation}\label{eq:data_shap:montecarlo}
    \phi_i = \E_{\pi\sim\Pi}[V(\setS^i_{\pi} \cup \{i\}) - V(\setS^i_{\pi})]
\end{equation}
where $S^i_{\pi}$ denotes the first $i-1$ samples in the training data set $\setS$ under the permutation $\pi$. Furthermore, $\setS^i_{\pi} \cup \{i\}$ denotes the addition of the $i$-th training data sample to the $S^i_{\pi}$.
In order to completely avoid the computationally expensive refitting of $\classifier(\cdot)$ in~\refeq{eq:data_shap:montecarlo}, the same authors~\cite{ghorbani2019data} propose to only perform a single gradient descent step instead of completely refitting $\classifier(\cdot)$ in~\refeq{eq:data_shap:montecarlo} -- i.e. the influence scores $\phi_i$ are estimated "on the fly" while training the model $\classifier(\cdot)$.

\subsection{Counterfactuals for Computational Recourse}\label{sec:counterfactuals}
A counterfactual explanation (often just called counterfactual) suggests changes to an input's features to alter the system's output, often requested for unfavorable outcomes~\cite{riveiro2022challenges} as a form of (computational) \textit{recourse}~\cite{karimi2021survey}. They are popular~\cite{CounterfactualReviewChallenges} because they mimic human reasoning in explanations~\cite{CounterfactualsHumanReasoning}.

The computation of a classic counterfactual~\cite{CounterfactualWachter} $\deltacf \in \RN^\dimsym$ for a given instance $\xorig \in \RN^\dimsym$ is phrased as the following optimization problem:
\begin{equation}\label{eq:counterfactualoptproblem}
\underset{\deltacf \,\in\, \RN^\dimsym}{\arg\min}\; \loss\big(\classifier(\xorig 
+ \deltacf), \ycf\big) + C \cdot \regularization(\deltacf)
\end{equation}
where $\loss(\cdot)$ penalizes deviation of the prediction $\classifier(\xorig + \deltacf)$ from the requested outcome $\ycf$, $\regularization(\cdot)$ states the cost of the explanation (e.g.\ cost of recourse) which should be minimized, and $C>0$ denotes the regularization strength balancing those two objectives.
The short-hand notation $\myCF{\x}{\classifier}$ denotes the solution to~\refeq{eq:counterfactualoptproblem} iff the target outcome $\ycf$ is uniquely determined.
Note that the cost $\regularization(\cdot)$ is domain-specific, but many implementations default to using the p-norm~\cite{guidotti2022counterfactual}.
\begin{remark}\label{remark:costrecourse}
    In the case of recourse -- i.e.\ turning an unfavorable into a favorable outcome --, we refer to the cost $\regularization(\deltacf)$, as the \emph{cost of recourse}.
\end{remark}
In this work, w.l.o.g., we refer to $y=0$ as the unfavorable, and $y=1$ as the favorable outcome.
Besides those two essential objectives in~\refeq{eq:counterfactualoptproblem}, there exist additional relevant aspects such as plausibility~\cite{CounterfactualGuidedByPrototypes,poyiadzi2020face}, diversity~\cite{mothilal2020explaining}, robustness~\cite{artelt2021evaluating,jiang2024robust}, etc.\ which have been addressed in literature~\cite{guidotti2022counterfactual}. However, the basic formalization~\refeq{eq:counterfactualoptproblem} is still very popular and widely used in practice~\cite{guidotti2022counterfactual,CounterfactualReviewChallenges}.

A critical and still unsolved fairness issue in computational recourse is the difference in the cost of recourse $\regularization(\deltacf)$ between protected groups~\cite{artelt2023ijcai,sharma2021fair,von2022fairness} -- i.e.\ individuals from one protected group (e.g.\ gender) get more costly recommendations on how to achieve recourse. It was shown that such cases could be created intentionally by targeted attacks~\cite{artelt2024effect,slack2021counterfactual}.

\subsubsection{Implementation}
There exist numerous methods and implementations/toolboxes for computing counterfactual explanations in practice~\cite{guidotti2022counterfactual} --  most include some additional aspects such as plausibility, diversity, etc.
\textit{Counterfactuals Guided by Prototypes}~\cite{CounterfactualGuidedByPrototypes} focuses on plausibility. Here a set of plausible instances (so-called prototypes) are used to pull the final counterfactual instance $\xcf$ (i.e.\ $\xcf := \xorig + \deltacf$) closer to these plausible instances and by this make them more plausible.
The \textit{Nearest Unlike Neighbor method} is a straightforward baseline method for computing plausible counterfactuals by picking the closest sample, with the requested output $\ycf$, from a given set (e.g.\ training set) as the counterfactual instance $\xcf$. 

\section{Influence of Training Samples on Explanations}\label{sec:influential_samples}
We consider scenarios, where a predictive model $\classifier_{\setD_\text{train}}: \setX\to\setY$ is derived from a given training data set $\setD_\text{train}$. Additionally, we assume an explanation generation mechanism $\text{expl}(\classifier_{\setD_\text{train}}, X)$ that produces either a local or global explanation $\expl$ within the set $\in\setExpl$ for the given $\classifier_{\setD_\text{train}}(\cdot)$. In the case of a local explanation, $X$ represents the sample or region for which an explanation is computed; otherwise, $X$ is ignored.
Note that we do not make any assumption on $\text{expl}(\cdot, \cdot)$ -- it could be any explanation or a related metric, such as the cost of recourse in the context of counterfactual explanations.

\emph{In this work, we are interested in identifying training samples in $\setD_\text{train}$ that had a high influence on an observed explanation $\text{expl}(\classifier_{\setD_\text{train}}, X)$ -- e.g.\ outlier or malicious training samples that shaped the observed explanation.}

\subsection{Quantifying the Influence of Training Samples on Explanations}
To quantify the impact of training samples on a given explanation, we need a metric for evaluating the similarity between explanations -- i.e. a mechanism that evaluates how similar or different two given explanations are.
For this purpose, we introduce a function $\Psi: \setExpl \times \setExpl \to \RN$ that computes the similarity of two given explanations $\expl_1, \expl_2\in \setExpl$ -- where we assume the same explanation generation mechanism $\text{expl}(\classifier, X)$ for $\expl_1$ and $\expl_2$ but not necessarily the same training data set from which $\classifier(\cdot)$ was derived.
We require that $\Psi(\expl_1, \expl_2) = 0 \leftrightarrow \expl_1 = \expl_2$; other than that, the computed real number is supposed to indicate their difference -- i.e. a larger output corresponds to a larger difference of the two given explanations $\expl_1, \expl_2$. Note, that the sign of the output of $\Psi(\cdot, \cdot)$ may offer additional insights, such as indicating direction or comparing magnitudes -- e.g. comparing the cost of recourse of two counterfactual explanations.
\emph{Example: In the case of explanations that are stated as real-valued vectors (i.e.\ $\setExpl=\RN^m$), $\Psi(\cdot)$ could be implemented by comparing their lengths: $\Psi(\expl_1, \expl_2) = \pnorm{\expl_1}_p - \pnorm{\expl_2}_p$, i.e. comparing the recourse costs of the counterfactuals.}

Based on $\Psi(\cdot, \cdot)$ we characterize influential training samples as follows:
\begin{definition}[Influential Training Samples]\label{def:influential_training_sample}
For a given training set $\setD_\text{train}$, we say that $\setD_\text{infl}\subset\setD_\text{train}$ has a strong influence on the explanation $\text{expl}(\classifier_{\setD_\text{train}}, X)$ iff the absence of $\setD_\text{infl}$ changes the explanation $\text{expl}(\classifier_{\setD_\text{train}}, X)$ significantly, i.e.:
\begin{equation}\label{eq:influential_training_sample}
    \Big|\Psi\big(\mathbb{E}[\text{expl}(\classifier_{\setD_\text{train}}, X)], \mathbb{E}[\text{expl}(\classifier_{\setD_\text{train}\setminus\setD_\text{infl}}, X)]\big)\Big| \gg 0
\end{equation}
\end{definition}
Note that the expected value $\mathbb{E}[\cdot]$ in~\refeq{eq:influential_training_sample} is accounting for randomness in the training process of $\classifier_{\setD}(\cdot)$ which for instance naturally occurs in the training of neural networks. %but also for some sampling-based explanation methods such as LIME.

\subsection{Reduction to a Game-theoretic Approach}
In this work, we propose to find such influential training samples (\refdef{def:influential_training_sample}) by conceptualizing a game-theoretic approach, namely the Data-SHAP method~\cite{ghorbani2019data}.

For this, we assume that $\Psi(\expl_1, \expl_2)$ can be decomposed as $\Psi(\expl_1, \expl_2) = V(\expl_1) - V(\expl_2)$ for some $V:\setExpl\to\RN$.
%\emph{In the context of our running example of counterfactuals, the decomposition of $\Psi(\expl_1, \expl_2)$ occurs naturally as $V(\setD \cup \{i\}) = \pnorm{\expl_{\setD\cup\{i\}}}_p$ and $V(\setD) = \pnorm{\expl_{\setD}}_p$.}
This assumption allows us to use the Data-SHAP method~\cite{ghorbani2019data} (\refeq{eq:data_shape}) for quantifying the influence $\phi_i$ of a single training sample:
\begin{equation}\label{eq:final_value}
   V(\setD \cup \{i\}) - V(\setD) := \Psi\big(\text{expl}(\classifier_{\setD \cup \{i\}}, X), \text{expl}(\classifier_{\setD}, X)\big)
\end{equation}

Given the computational complexity of~\refeq{eq:data_shape} and the potentially complex computation of explanations in~\refeq{eq:final_value}, we propose to use a gradient-based Monte-Carlo approach similar to the one in the original Data-SHAP paper~\cite{ghorbani2019data} -- assuming that the predictive model $\classifier(\cdot)$ is differentiable and can be trained using gradient-descent. Here, we estimate the $\phi_i$ scores while training the model $\classifier(\cdot)$ -- i.e. instead of training the model $\classifier(\cdot)$ to convergence, we already evaluate the explanations at training time.
The complete gradient-based Monte-Carlo algorithm for computing the influence score $\phi_i$ of each training sample on an explanation is described in Algorithm~\ref{algo:scoring}.
\begin{algorithm}[t!]
\caption{Finding Influential Training Samples}\label{algo:scoring}
\textbf{Input:} Labeled training set $\setD$; $K$ number of training steps; Differentiable model $\classifier(\cdot)$; Single or group of samples $X$ which are going to be explained\\
\textbf{Output:} Influence score $\phi_i$ of each training sample.
\begin{algorithmic}[1]
  \For{$j=1, ..., N$ or until convergence}
    \State Random initialization of $\classifier(\cdot)$ weights $\w$
    \For{$t=1, ..., K$} \Comment{Training loop}
        \State $\expl_{\setD} = \text{expl}(\classifier, X)$  \Comment{Initial explanation}
        \State $\pi = \text{random\_permutation}(|\setD|)$
        \For{idx $\in \pi$}
            \State $(\x_i, \y_i) = \setD\text{[idx]}$ \Comment{Get current sample}
            \State $\w -= \lambda\nabla_{\w}\loss(\x_i, \y_i \mid \w)$ \Comment{Gradient descent}
            \State $\expl_{\setD\cup\{i\}}= \text{expl}(\classifier, X)$  \Comment{Compute new explanation}
            \State $\phi_\text{[idx]}^{j,t} = \Psi(\expl_{\setD\cup\{i\}}, \expl_{\setD})$ \Comment{Compute influence of $(\x_i, \y_i)$}
            \State $\expl_{\setD} = \expl_{\setD\cup\{i\}}$ \Comment{Update current explanation}
        \EndFor
    \EndFor
  \EndFor
  \State $\phi_i = \frac{1}{N \cdot K}\sum_{j,t} \phi_i^{j,t}$  \Comment{Final influence scores}
\end{algorithmic}
\end{algorithm}

The core part of Algorithm~\ref{algo:scoring} are lines 3-11.
The neural network (or any other differentiable model $\classifier(\cdot)$) is initialized with random parameters $\w$ and trained for $K$ iterations. In each iteration, the entire training data set $\setD$ is shuffled (line 5) and then each training sample $(\x_i, \y_i)$ is processed as follows: 1) update the model parameters $\w$ using gradient descent (line 8); 2) Recompute the explanation $\text{expl}(\classifier, X)$ of interest (line 9); 3) Compare current explanation with the explanation from the previous iteration (line 10) -- this allows us to estimate (\refeq{eq:final_value}) the influence of the current training sample $(\x_i, \y_i)$ on the explanation of interest.
This procedure is repeated several times until the influence scores $\phi_i$ converge.
Then, we select the training samples with the largest absolute influence score as the set of most influential training samples $\setD_\text{infl}$ (\refdef{def:influential_training_sample}):
\begin{equation}
\begin{split}
    \setD_\text{infl} &= \{(\x_{i_0}, y_{i_0}), (\x_{i_1}, y_{i_1}), \dots\} \text{ s.t. } |\phi_{i_0}| \geq |\phi_{i_1}| \geq \dots
\end{split}
\end{equation}
where the number of influential training samples could either be given by the user or determined automatically by applying a minimum threshold to $|\phi_i|$.

\paragraph{Computational Considerations}
Since Algorithm~\ref{algo:scoring} likely requires many iterations to converge, the computation of the explanation $\text{expl}(\classifier, X)$ in each step might become a (computational) bottleneck. Therefore, approximations of the explanation generation mechanism $\text{expl}(\classifier, X)$ might be necessary as proposed in the two case studies.

\section{Case-Study I: Cost of Recourse}\label{sec:casestudyi}
Here, we consider the challenge of identifying training samples (see Algorithm~\ref{algo:scoring}) that significantly impact the average cost of recourse -- i.e., the average cost it takes to change the outcome for negatively (i.e. unfavorable) classified instances.
We aim to identify those training samples that have an increasing effect on the average cost of recourse -- those training samples might be suitable candidates for manual inspection and (potential) deletion in order to decrease the average cost of recourse.

Assuming that we have a set of negatively classified instances $\setD$, our quantity of interest, the average cost of recourse is defined as follows:
\begin{equation}\label{eq:avg_cost_recourse}
    \text{expl}(\classifier_{\set{S}}, \setD) = \frac{1}{|\setD|}\sum_{\x_i \in \setD} \regularization\circ\myCF{\x_i}{\classifier_{\set{S}}}
\end{equation}
Note that the training data set $\set{S}$ and the (test) set $\setD$ are not necessarily the same -- i.e.\ one might be interested in evaluating on a hold-out data set $\setD$ that is disjunct from the training data set $\set{S}$.

However, the evaluation of~\refeq{eq:avg_cost_recourse} is computationally expensive because it requires computing a counterfactual $\myCF{\x_i}{\classifier_{\set{S}}}$ for every sample in $\setD$.
In order to make the computation of the influence scores $\phi_i$ feasible, we assume that the cost of recourse $\regularization(\cdot)$ is related to the distance to the decision boundary and propose an approximation based on the difference in the logits~\cite{sharma2021fair}:
\begin{equation}\label{eq:cfcost_approx}
    \regularization\circ\myCF{\x_i}{\classifier_{\set{S}})} \approx |g_0(\x_i) - g_1(\x_i)|
\end{equation}
where $g_1, g_2$ denote the logits of the neural network $\classifier_{\set{S}}(\cdot)$.
The authors of~\cite{sharma2021fair} show that~\refeq{eq:cfcost_approx} constitutes a good approximation of the distance to the decision boundary. We make use of it as a proxy for the cost of recourse.

The final value function $V(\setS)$ is then given as:
\begin{equation}\label{eq:casestudyi}
        V(\setS) = \frac{1}{N_{\setS}}\sum_{\x_i \in \setD \mid \classifier_{\set{S}}(\x_i)=0} \Big|g_0(\x_i) - g_1(\x_i)\Big| \quad N_{\setS} := |\{\x_i \in \setD \mid \classifier_{\set{S}}(\x_i)=0\}|
\end{equation}

We can then substitute~\refeq{eq:casestudyi} for~\refeq{eq:final_value} in Algorithm~\ref{algo:scoring} for computing the influence of each training sample.
The training samples of interest are those with a large positive influence score, i.e.\ $\phi_i \gg 0$, since those are increasing the average cost of recourse~\refeq{eq:avg_cost_recourse} (for negatively classified instances) when added to the training set. In the empirical evaluation (Section~\ref{sec:experiments}), we investigate the effect of removing those training samples from the training data.

\section{Case-Study II: Difference in the Cost of Recourse}\label{sec:method:diff_cost_recourse}
We consider the problem of analyzing the reasons for differences in the cost of recourse between two protected groups (here denoted by a binary attribute $q\in\{0, 1\}$), a significant fairness issue in computational recourse~\cite{artelt2023ijcai,sharma2020certifai}. Our goal is to identify training samples that increase this cost difference. These samples could be candidates for manual inspection and potential removal to reduce the disparity and enhance fairness in computational recourse.
Here, we consider the difference in the worst-case cost of recourse between two protected groups as our quantity of interest, which is formalized as follows:
\begin{equation}\label{eq:diff_cost_recourse}
    \text{expl}(\classifier_{\set{S}}, \setD) = \left|\underset{\x_i\in\setD}{\max}(\regularization\circ\myCF{\x_i \mid q=0}{\classifier_{\set{S}}}) - \underset{\x_i\in\setD}{\max}(\regularization\circ\myCF{\x_i \mid q=1}{\classifier_{\set{S}})}\right|
\end{equation}
where we only consider negatively classified samples in the given set $\set{D}$ -- we drop the explicit constraint $\classifier_{\set{S}}(\x_i) = 0$ for better readability.
%Consequently, we define the value function $V(\setS)$ to be equal to~\refeq{eq:diff_cost_recourse}.
%
As in the first case study, we achieve computational feasibility by approximating the cost of recourse by the difference in the logits:
\begin{equation}
    \regularization\circ\myCF{\x_i \mid q=?}{\classifier_{\set{S}})} \approx |g_0(\x_i) - g_1(\x_i)|
\end{equation}
The final value function $V(\setS)$ is then given as:
\begin{equation}\label{eq:final_valuefunction}
\begin{split}
        V(\setS) = \Big| &\underset{\x_i \in \setD \mid q=0}{\max}(|g_0(\x_i) - g_1(\x_i)|) -\underset{\x_i \in \setD \mid q=1}{\max}(|g_0(\x_i) - g_1(\x_i)|)\Big|
\end{split}
\end{equation}

We can then substitute~\refeq{eq:final_valuefunction} for~\refeq{eq:final_value} in Algorithm~\ref{algo:scoring} for computing the influence of each training sample.
Again, the training samples of interest to us are those with a large positive influence score, i.e.\ $\phi_i \gg 0$.
In the empirical evaluation (Section~\ref{sec:experiments}), we investigate the effect of removing those training samples from the training data set.

\section{Experiments}\label{sec:experiments}
We seek to empirically answer the following two research questions:
\begin{enumerate}[label=RQ\arabic*]
    \item\label{rq1} Correctness of our proposed Algorithm~\ref{algo:scoring}: Evaluate how the deletion of those identified training samples influences the counterfactuals, as well as the predictive performance (F1-score) of the classifier.
    \item\label{rq2} Improvement of our proposed Algorithm~\ref{algo:scoring} over existing data valuation method: Are highly influential training samples on explanations (\refdef{def:influential_training_sample}) different from those for (only) influencing the predictive performance?
\end{enumerate}

\subsection{Data}
We consider the following two benchmark data sets from the fairness literature~\cite{friedler2019comparative}: 1) The ``Diabetes'' data set~\cite{efron2004least} (denoted \textit{Diabetes}) contains data from $442$ diabetes patients, each described by $9$ numeric attributes together with the sensitive binary attribute ``sex''. The target is a binarized quantitative measure of disease progression one year after baseline.
; 2) The ``German Credit Data set''~\cite{GermanCreditDataSet} (denoted \textit{Credit}) is a data set for loan approval and contains $1000$ samples each annotated with $7$ numerical and $13$ categorical attributes, including the sensitive binary attribute ``sex'', 
with a binary target value. We only use the seven numerical features.

\subsection{Setup}
We use the $\ell_1$ norm as a popular implementation~\cite{guidotti2022counterfactual} of the cost of recourse -- i.e.\ $\regularization(\cdot)=\pnorm{\cdot}_1$ -- and utilize a Multi-Layer Perceptron (with two hidden layers) as the classifier $\classifier(\cdot)$.

We conduct the experiments in a $5$-fold cross-validation: 1) fitting the classifier; 2) computing the most influential training samples, whereby all negative classified samples (i.e.\ $\classifier(\cdot)=0$) from the test set are considered. % when computing the computational recourse.
We compute the computational recourse (i.e.\ counterfactual) on the test set using three different popular recourse methods:
Nearest Unlike Neighbor (denoted \textit{NUN}), as a simple baseline for plausible counterfactuals;
Counterfactuals guided by Prototypes~\cite{CounterfactualGuidedByPrototypes} (denoted \textit{Proto}) as an advanced method for computing plausible counterfactuals;
Classic counterfactuals~\cite{CounterfactualWachter} (denoted \textit{Wachter}) by solving~\refeq{eq:counterfactualoptproblem}.

For the purpose of investigating \ref{rq1}, we increasingly remove (1\% - 30\%) of the most influential training samples from the training set, retrain the classifier, and re-evaluate the quantity of interest on the (unchanged) test set -- i.e. either the average cost of recourse (case study I) or the difference in the cost of recourse (case study II).
To account for the randomness in training the neural network classifier, we take the average (expectation) of the quantity of interest over five training runs.

\textit{Baselines}
For the purpose of investigating \ref{rq2}, we compare our method (i.e.\ Algorithm~\ref{algo:scoring}) to two baselines:
1) Removal of random training samples;
2) Removal of the most influential training samples on the predictive performance as returned by the original Data-SHAP method~\cite{ghorbani2019data}.

\subsection{Case Study I}\label{sec:experiments:casestudyi}
\begin{figure*}[t!]
    \centering
    \includegraphics[width=0.24\textwidth]{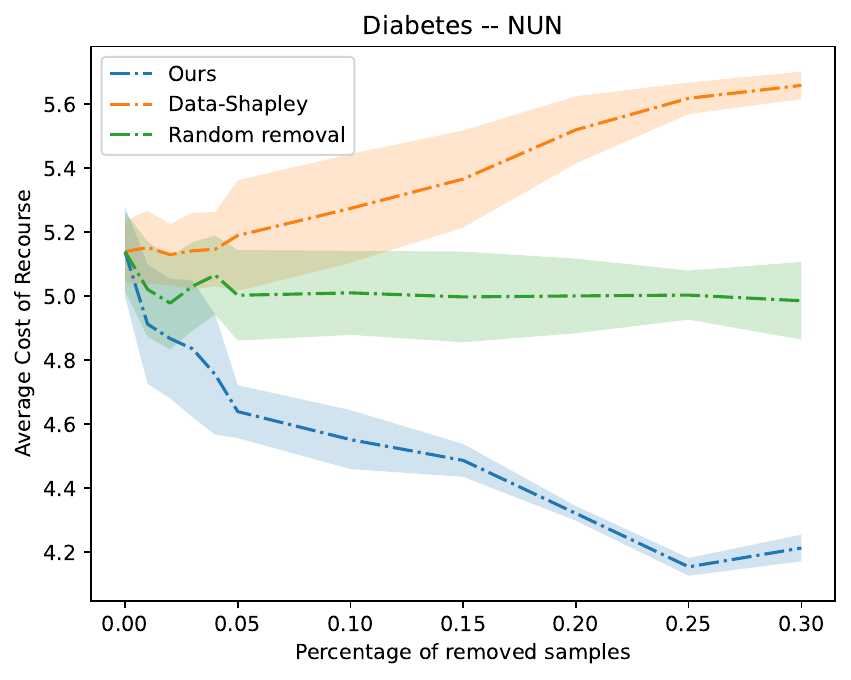}
    \includegraphics[width=0.24\textwidth]{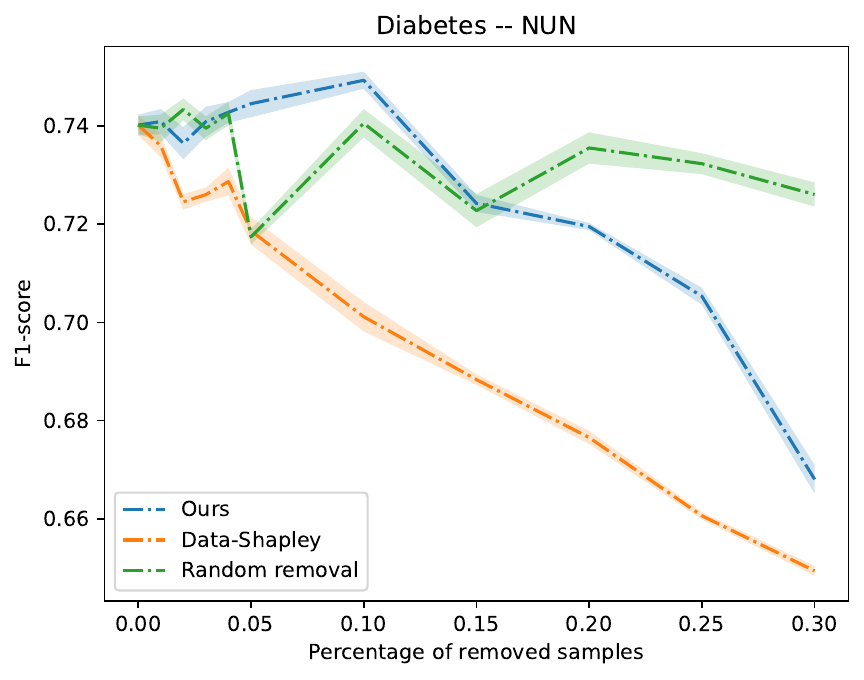}
    \includegraphics[width=0.24\textwidth]{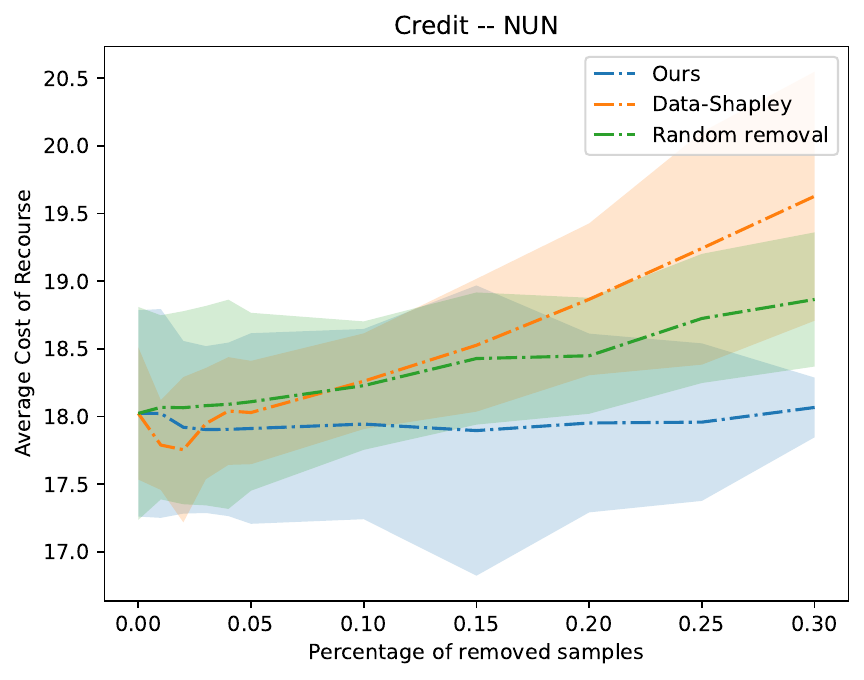}
    \includegraphics[width=0.24\textwidth]{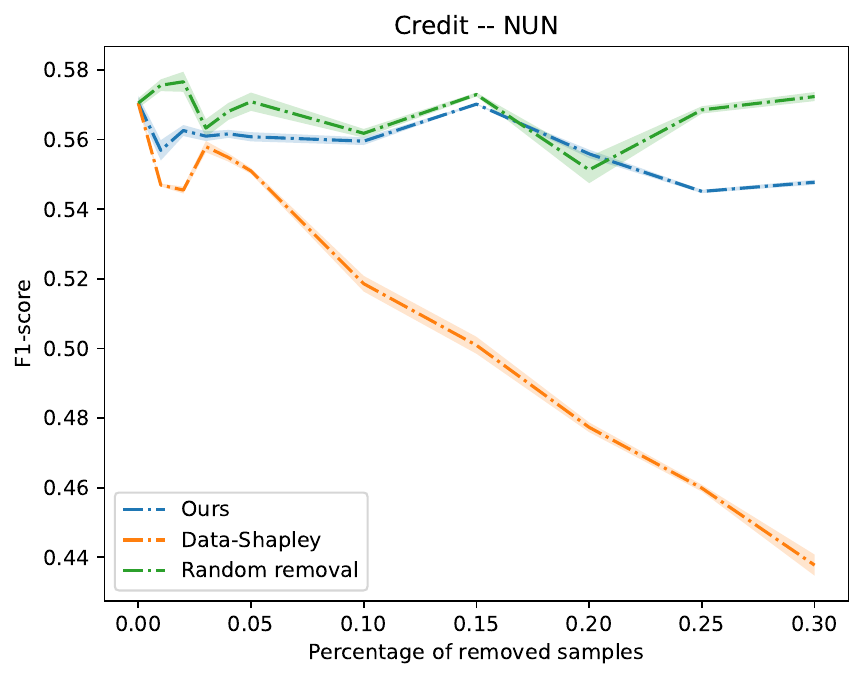}

    \includegraphics[width=0.24\textwidth]{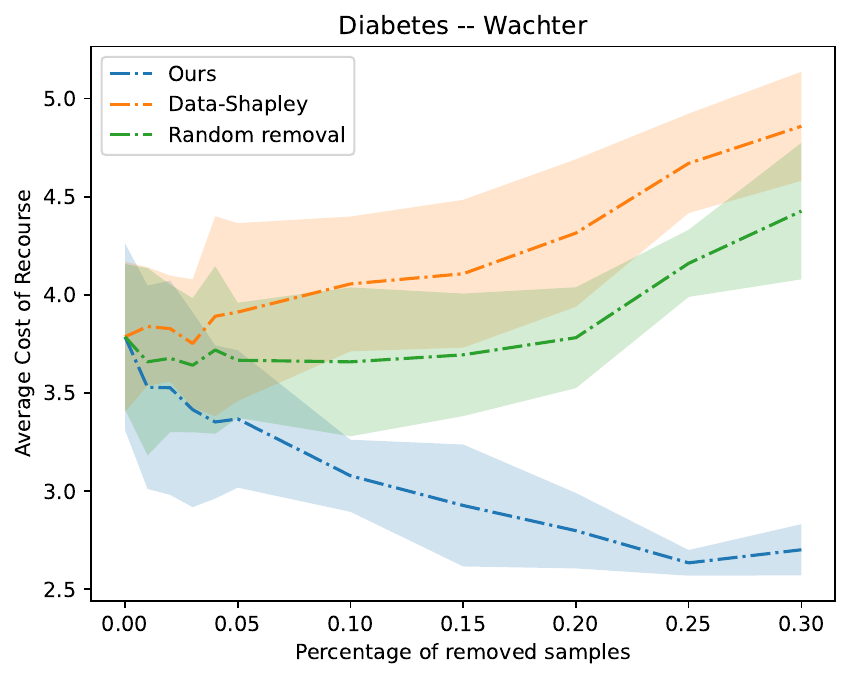}
    \includegraphics[width=0.24\textwidth]{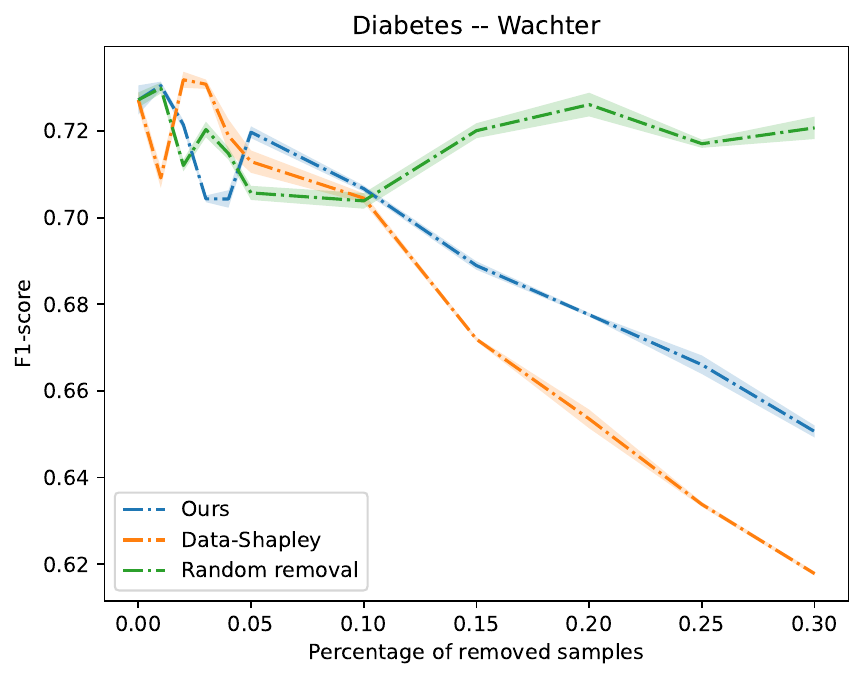}
    \includegraphics[width=0.24\textwidth]{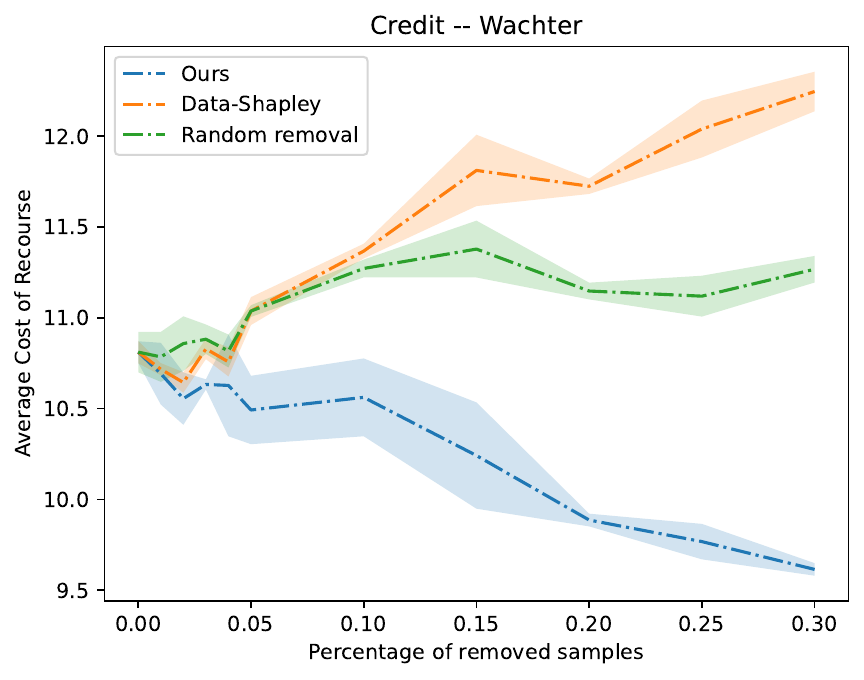}
    \includegraphics[width=0.24\textwidth]{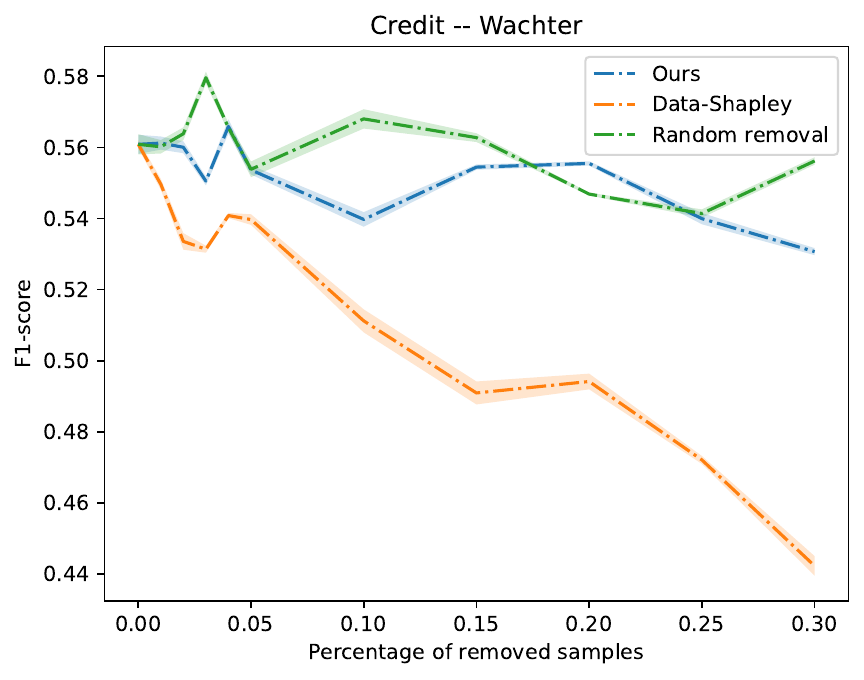}

    \includegraphics[width=0.24\textwidth]{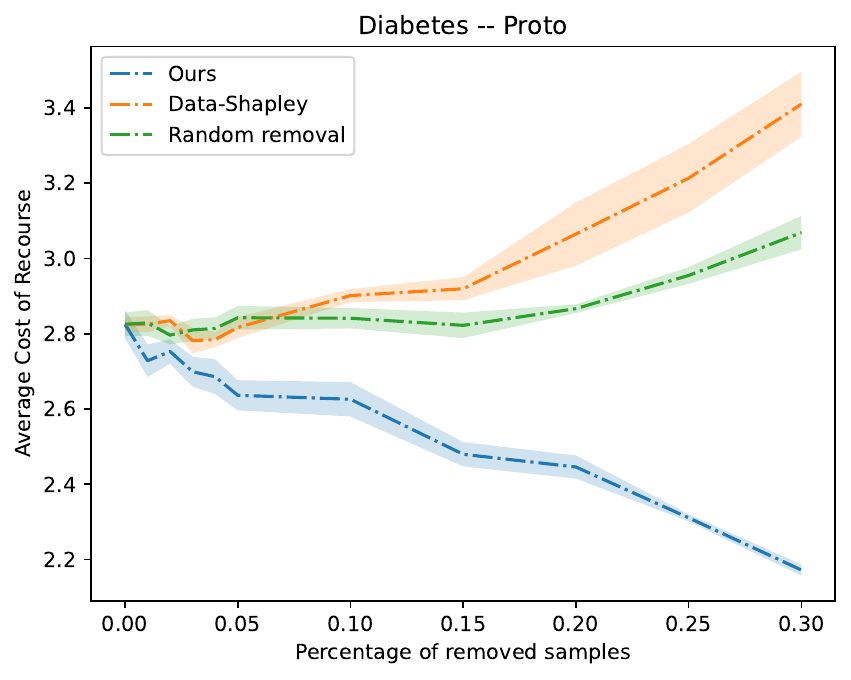}
    \includegraphics[width=0.24\textwidth]{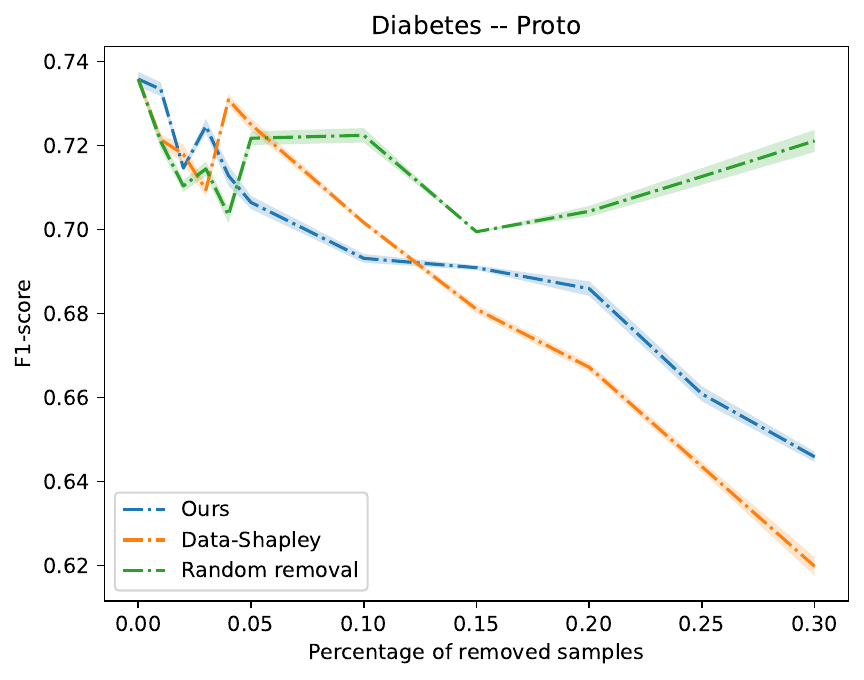}
    \includegraphics[width=0.24\textwidth]{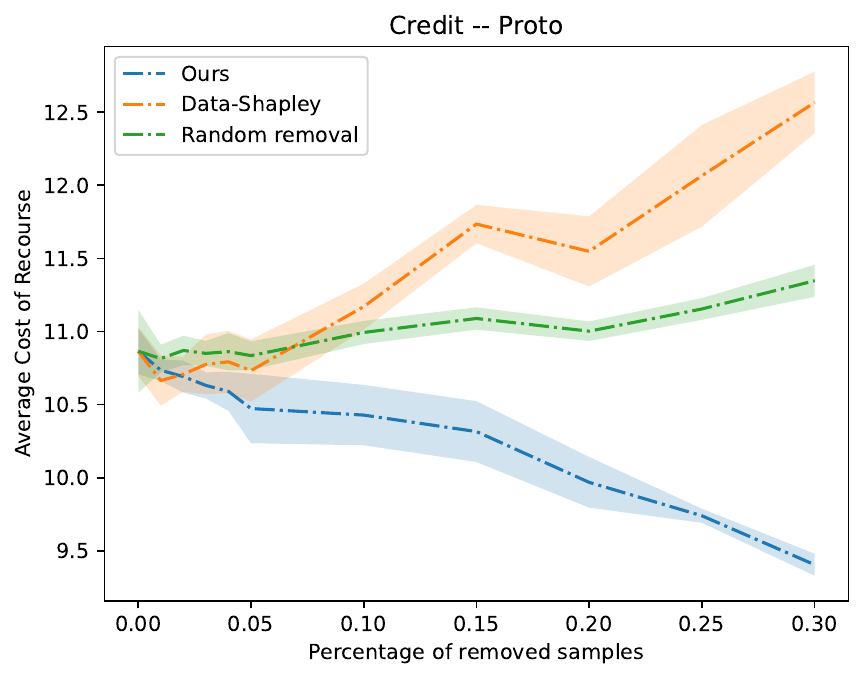}
    \includegraphics[width=0.24\textwidth]{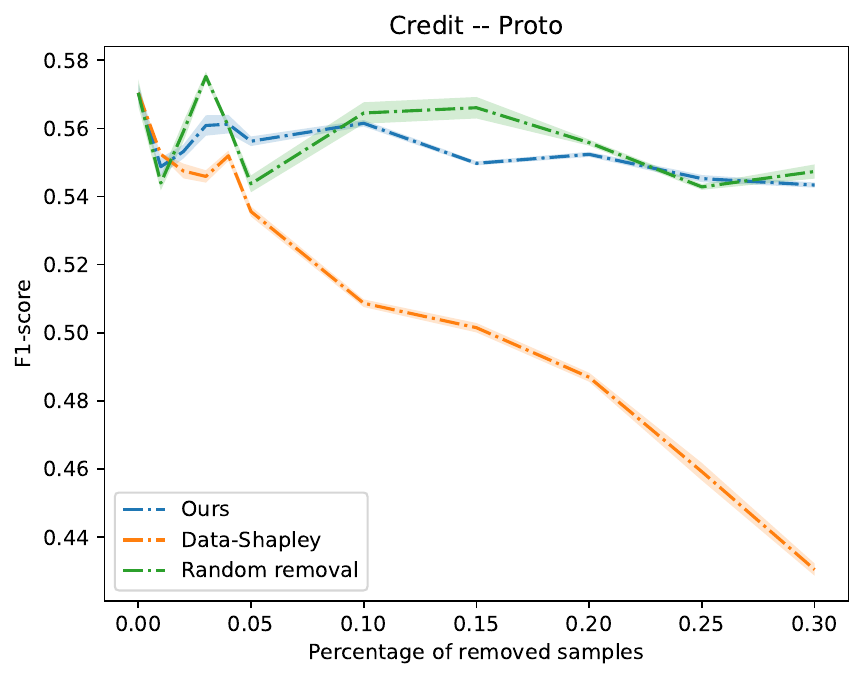}
    
    \caption{Case-Study I: Effect of removing training samples that have a high influence on the average cost of recourse~\refeq{eq:avg_cost_recourse} -- we show (mean \& variance over all folds) the effect on the average cost of recourse, as well as on the predictive performance (i.e. F1-score).}
    \label{fig:results:globalcostrecourse}
\end{figure*}
\begin{figure*}[t!]
    \centering

    \includegraphics[width=0.24\textwidth]{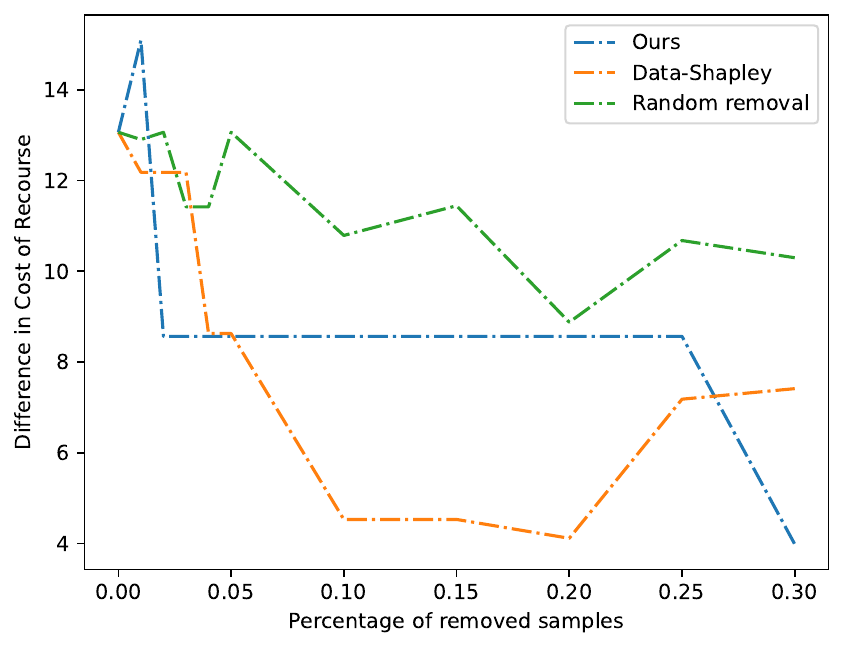}
    \includegraphics[width=0.24\textwidth]{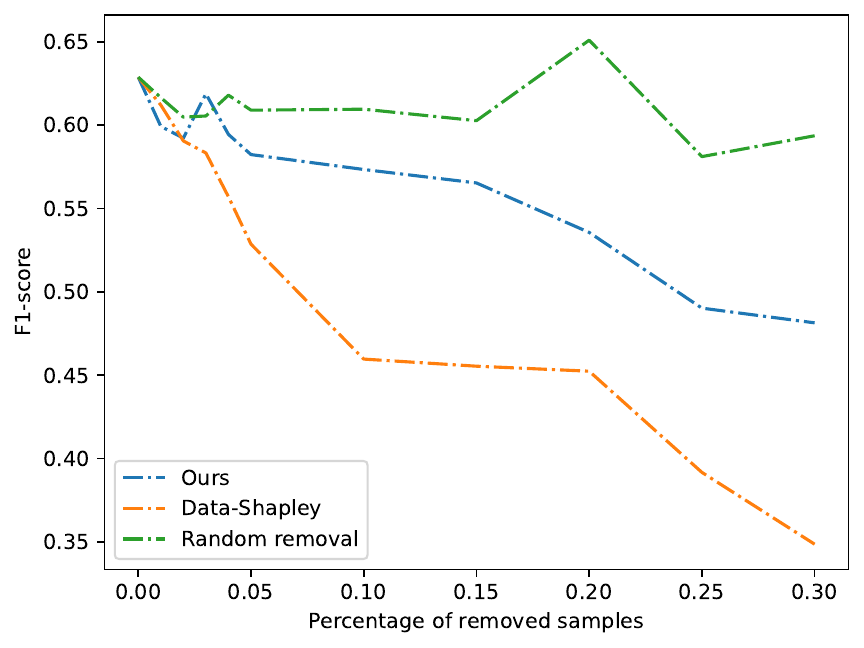}
    \includegraphics[width=0.24\textwidth]{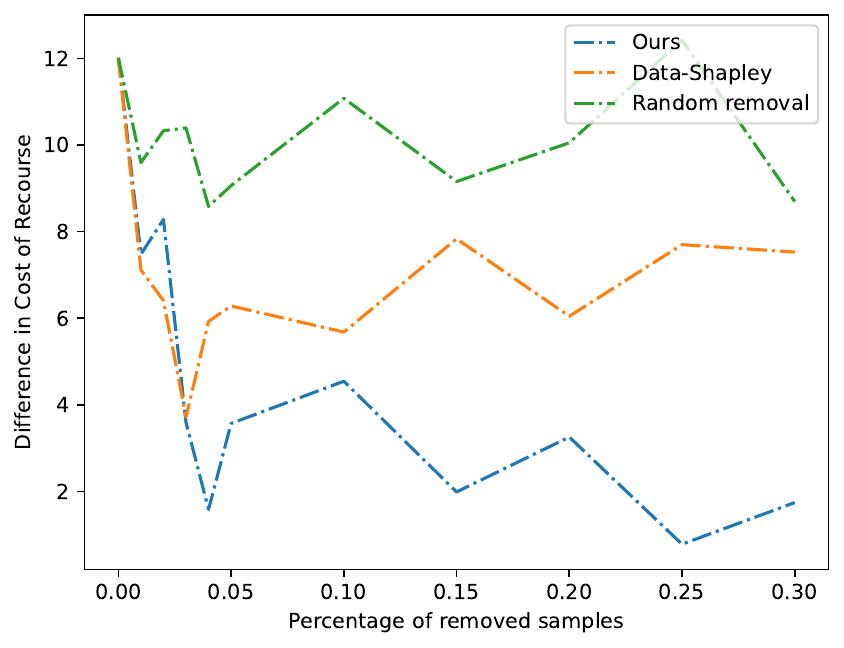}
    \includegraphics[width=0.24\textwidth]{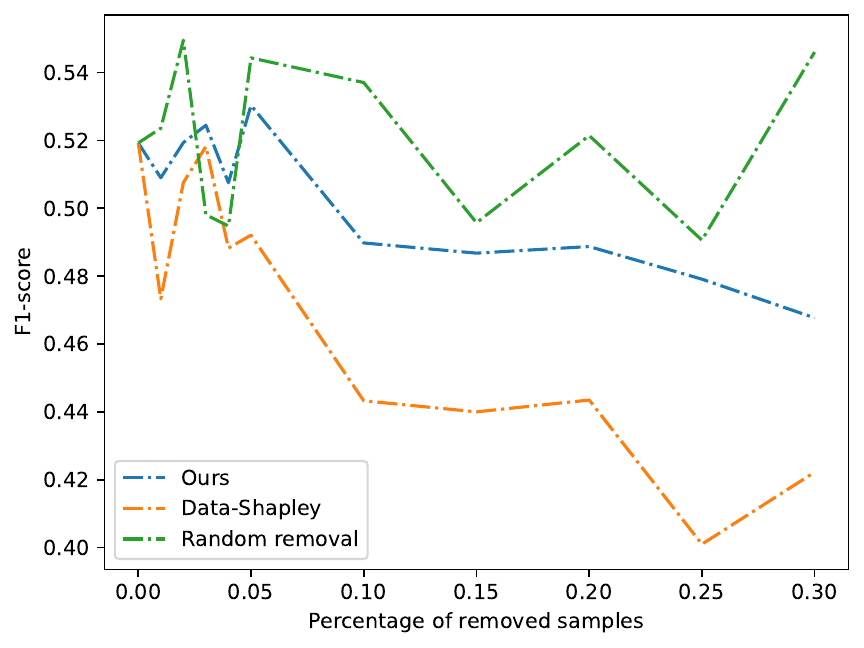}

    \caption{Case Study II: Effect of removing training samples that have a high influence on the difference in the cost of recourse~\refeq{eq:diff_cost_recourse} -- we show the effect on the difference in the cost of recourse, as well as on the predictive performance (i.e. F1-score).
    Note that we only consider the train-test split with the worst original difference.}
    \label{fig:results:groupfair}
\end{figure*}
We consider all possible combinations of the two aforementioned data sets and the three aforementioned counterfactual explanation methods. In Figure~\ref{fig:results:globalcostrecourse}, we report the mean as well as the variance of all measurements. Note that for the average cost of recourse, smaller numbers are better, while for the F1-score, larger numbers are better.

\ref{rq1}) We observe that in almost all cases, our method is able to identify influential training samples (\refdef{def:influential_training_sample}) that, if removed, indeed decrease the average cost of recourse significantly.
\ref{rq2}) The two baselines (random removal and Data-SHAP~\cite{ghorbani2019data}) return training samples that, if removed, often increase (instead of decrease) the average cost of recourse.
This shows that training samples reducing the average cost of recourse differ from those positively affecting the model's predictive performance (as identified by Data-SHAP). Thus, existing data valuation methods may be insufficient, stressing the necessity of specialized methods like our proposed Algorithm~\ref{algo:scoring} for identifying training samples that shaped a given explanation.

We observe that random removal typically has little effect, whereas our method and Data-SHAP negatively impact predictive performance -- which is to be expected given the fact that (potentially useful) information is removed from the training data. For Data-SHAP, this result aligns with its design and the original paper's findings~\cite{ghorbani2019data}. Although our method also reduces predictive performance, the drop is less severe than with Data-SHAP.
Finally, all effects are getting stronger the more training samples are removed.

\subsection{Case Study II}\label{sec:experiments:casestudyii}
\begin{table}[t!]
 \caption{Case Study II: Differences in the cost of recourse between the protected groups~\refeq{eq:diff_cost_recourse} for the Credit data set. We report maximum and average \& variance.}
 \label{table:exp:diff_cost_recourse:scores}
\centering
\begin{tabular}{||c|c||c|c||c|c||c|c||c|c|c|c||}
 \hline
 \multicolumn{2}{||c||}{NUN}  & \multicolumn{2}{|c||}{Proto} & \multicolumn{2}{|c||}{Wachter}  \\
 \hline
 Max. & Avg. $\pm$ Var.  & Max. & Avg. $\pm$ Var. & Max. & Avg. $\pm$ Var.\\
 \hline\hline
  $13.06$ & $5.77 \pm 4.43$ & $12.01$ & $8.31 \pm 2.20$ & $2.97$ & $2.07 \pm 0.53$  \\
  \hline
 \end{tabular}
\end{table}
We focus on the German Credit dataset~\cite{GermanCreditDataSet} due to its known fairness issues in certain train-test splits~\cite{friedler2019comparative}. We assess the difference in recourse costs between protected groups for each split and counterfactual explanation method, as shown in Table~\ref{table:exp:diff_cost_recourse:scores}. Significant differences are found for the Nearest Unlike Neighbor (NUN) and prototype-guided counterfactuals, while Wachter's method~\cite{CounterfactualWachter} shows no significant difference. High variance is observed because not all splits exhibit unfairness. We evaluate the three methods on the split with the highest unfairness in recourse cost differences. The results are shown in Figure~\ref{fig:results:groupfair}, where for the cost difference smaller numbers indicate better fairness, and larger numbers indicate better F1-scores.

\ref{rq1}) For decreasing the difference in the cost of recourse~\refeq{eq:diff_cost_recourse} %-- i.e. improving group fairness in computational recourse --
, we observe that our method almost always achieves better or at least competitive performance with the Data-SHAP method~\cite{ghorbani2019data}.
\ref{rq2}) However (similar to case study I), our method consistently maintains a much better predictive performance of the classifier than the Data-SHAP method which often leads to a drastic drop in predictive performance. Random removal of training samples does not affect the F1-score as much as the other two methods do, however, it also does not significantly decrease the difference in the cost of recourse.

Similar to our findings in the first case study, these results demonstrate that our proposed method is able to identify influential training samples (\refdef{def:influential_training_sample}) and that those samples do not affect the predictive performance as much as the Data-SHAP method.

\section{Summary \& Conclusion}
We introduced the novel problem of identifying training samples that have a strong impact on given explanations and proposed a Data-SHAP~\cite{ghorbani2019data}-based algorithm for this purpose. We explored two case studies: the cost of recourse (case study I) and the difference in recourse costs between two protected groups (case study II). Although we focused on the cost of recourse, our methodology and Algorithm~\ref{algo:scoring} are applicable to other explanations as well. Our empirical results show that removing identified samples significantly reduces recourse costs or group disparities without greatly harming predictive performance. Notably, when Data-SHAP~\cite{ghorbani2019data} competes with our method, it results in a much larger drop in predictive performance, indicating a fundamental difference between training samples affecting computational recourse and those affecting predictive performance.

It would be interesting to extend Algorithm~\ref{algo:scoring} to consider groups of training samples, as they might have a strong influence collectively. This presents the challenge of identifying interacting or dependent groups of samples. Additionally, while our algorithm performs well empirically, it lacks formal guarantees regarding the logit approximation and the effects of removing multiple samples. We leave these aspects for future work.

\textbf{Acknowledgments.}
This research was supported by the Ministry of Culture and Science NRW (Germany) as part of the Lamarr Fellow Network. This publication reflects the views of the authors only.

\bibliographystyle{splncs04}
\bibliography{aaai25}

\end{document}